%% file: main.tex
\documentclass[conference]{IEEEtran}
\usepackage{times}

\usepackage[numbers]{natbib}
\usepackage{amsmath}
\usepackage{multicol}
\usepackage{booktabs} 
\usepackage{amssymb}
\usepackage{multirow}
\usepackage{adjustbox}
\usepackage{graphicx}
\usepackage{svg}
\usepackage{caption}
\usepackage{float}
\usepackage[bookmarks=true]{hyperref}
\usepackage[font=small]{caption}
\graphicspath{{Images/}}
\usepackage{amssymb}
\usepackage{listings}

\usepackage{tcolorbox}
\tcbuselibrary{skins, breakable}

\makeatletter

\newcommand{\Rmnum}[1]{\expandafter\@slowromancap\romannumeral #1@}
\makeatother

\newtcolorbox{promptbox}[1]{
    colback=gray!5,          
    colframe=gray!50,        
    coltitle=black,          
    title={\textbf{#1}},     
    boxrule=0.5pt,          
    arc=1mm,                 
    left=1mm, right=1mm, top=1mm, bottom=1mm, 
    fontupper=\small\ttfamily, 
    breakable,              
    enhanced           
}

\begin{document}

\title{A Closed-Loop Multi-Agent Framework for Robust Multi-Robot Manipulation}

\twocolumn[{
\renewcommand\twocolumn[1][]{#1}

\author{\authorblockN{\textbf{Yi-Xiang He}\textsuperscript{1},
\textbf{Lan Wei}\textsuperscript{1},
\textbf{Haoming Cen}\textsuperscript{1},
\textbf{Jian-Jian Jiang}\textsuperscript{1},
\textbf{Zhuohao Li}\textsuperscript{1,5},\\
\textbf{Guanxing Lu}\textsuperscript{4},
\textbf{Yihan Yang}\textsuperscript{1},
\textbf{Dandan Zhang}\textsuperscript{3},
\textbf{Wei-Shi Zheng}\textsuperscript{1,2,5,$\dagger$}}

\authorblockA{\textsuperscript{1}Sun Yat-sen University,
\textsuperscript{2}Key Laboratory of Machine Intelligence and Advanced Computing, Ministry of Education, China\\
\textsuperscript{3}Imperial College London,
\textsuperscript{4}Tsinghua Shenzhen International Graduate School,
\textsuperscript{5}Shenzhen Loop Area Institute}
\authorblockA{\small $^\dagger$Corresponding author.}}

\vspace{0.5cm}
\maketitle

\begin{center}
    \vspace{-1cm}
    \captionsetup{type=figure}
    \includegraphics[width=\textwidth]{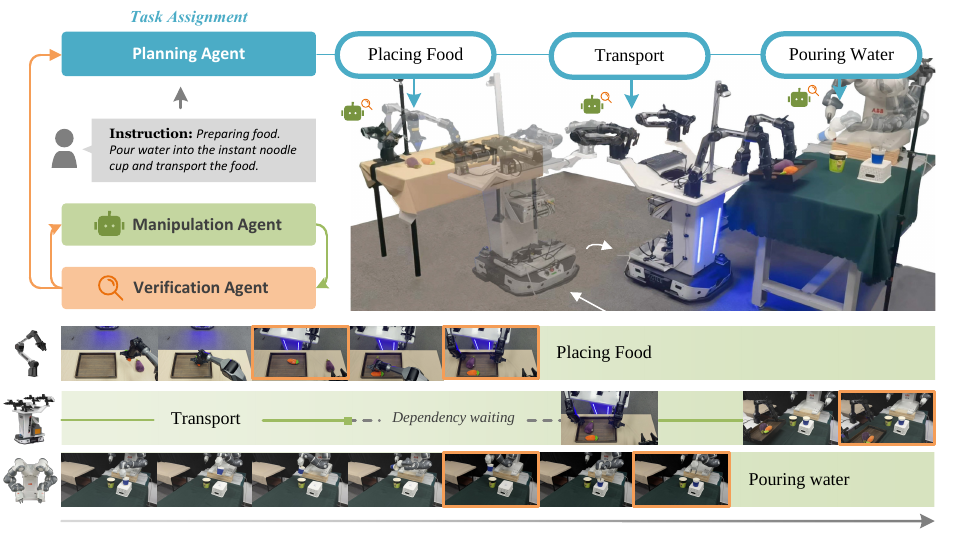}
    \captionof{figure}{We propose a closed-loop multi-agent framework for robust multi-robot manipulation. The system is driven by three specialized agents: \textbf{the Planning Agent} decomposes instructions into dependency-aware sub-tasks, \textbf{the Manipulation Agent} grounds these sub-tasks into physical actions through adaptive tool use, and \textbf{the Verification Agent} identifies execution errors to trigger hierarchical re-planning. Extensive real-world experiments demonstrate that our framework achieves versatile manipulation capabilities, effectively handles external disturbances and execution uncertainties, and adapts from single-robot operations to multi-robot collaborative tasks.}
    \vspace{0.3cm}
\end{center}
}]

\begin{abstract}
Multi-robot systems provide the parallelism and redundancy necessary for long-horizon tasks, while Large Language Models (LLMs) offer the reasoning capabilities to decompose these objectives into actionable plans. However, effectively grounding this high-level reasoning in physical multi-robot execution remains an open challenge. Existing LLM-based approaches fall mainly into two categories: Single-robot methods achieve robust contact-rich manipulation but lack the coordination mechanisms required for tasks spanning multiple workspaces. Current multi-robot frameworks focus on high-level planning, often treating manipulation as an idealized primitive that fails to account for real-world execution uncertainties. To address this, we propose a hierarchical closed-loop agentic LLM-based framework to ensure robust multi-robot manipulation. Our system consists of three specialized agents: the Planning Agent decomposes instructions into allocated sub-tasks, the Manipulation Agent for each robot executes actions via adaptive tool use, and the Verification Agent closes the loop by monitoring physical outcomes and feeding back semantic corrections. Extensive real-world experiments demonstrate that our framework achieves superior success rates, ensures robust adaptability ranging from single to cross workspace manipulation, and offers a generalizable approach for diverse manipulation tasks.
\end{abstract}

\IEEEpeerreviewmaketitle

\section{Introduction} 

Multi-robot systems demonstrate significant potential in diverse environments, from domestic service to industrial manufacturing \cite{housekeeping, factory, factory1}, by expanding manipulation capabilities and enhancing efficiency to offer great system extensibility. However, achieving such systems imposes a substantial burden on manipulation complexity, necessitating coordination between heterogeneous robots and effective emergency response to unexpected failures during execution.

Recently, the integration of Large Language Models (LLMs) has offered new avenues for robotic manipulation by leveraging their capabilities in semantic understanding and logical reasoning \cite{qwen3, gemini, gpt}. Existing LLM-based approaches fall broadly into two categories. The first uses LLMs to synthesize single-robot manipulation policies, code, grasping guidance, or constraints \cite{CaP, CoPa, Rekep, graspasyousay, afforddexgrasp, typetele}. While these methods demonstrate promising physical grounding, they lack the coordination mechanisms required for tasks that span multiple workspaces. The second paradigm targets multi-robot systems but focuses primarily on high-level task allocation and symbolic planning \cite{MR1, llmmr1, llmmrsurvey, Smartllm, coherent}, typically validating plans prior to execution while treating manipulation as an idealized primitive. This abstraction assumes deterministic success and neglects real-world execution uncertainties such as grasp failures, contact variability, external disturbances, or unreachable workspace.

To achieve robust and self-correction capabilities in real-world tasks, we propose a closed-loop agentic workflow for robust multi-robot manipulation. Our core insight is that tackling real-world uncertainties requires a shift from static execution pipelines to an iterative agentic process that incorporates a hierarchical recovery mechanism. Unlike previous methods that restrict validation to the pre-execution phase and subsequently rely on idealized actions, an agentic workflow empowers the system to actively reason, act, and reflect on its performance. By integrating semantic planning, physical grounding, and outcome verification, our architecture enables robots to perceive environmental feedback and adaptively recover from failures, achieving robust and verifiable multi-robot manipulation and ensuring reliability by allowing agents to actively address execution uncertainties in real-world environments. 

Specifically, we instantiate this workflow with three specialized agents that operate in a closed-loop cycle of reasoning, acting, and reflecting, enabling refinement of both high-level collaboration decisions and low-level execution. The information flow begins with the Planning Agent, which serves as the reasoning core to handle task decomposition and allocation while balancing semantic instruction logic with the specific capabilities of heterogeneous robots. To ensure accurate semantic grounding, it employs a bidirectional interaction mechanism to query bottom-up perception from the Manipulation Agent whenever instructions are ambiguous. Following this, the Manipulation Agent takes charge of the acting phase. It translates sub-tasks into fine-grained operations by adaptively leveraging tools such as keypoint localization and grasp generation models to ground semantic plans into reality. Finally, to tackle inevitable execution uncertainties, the verification functions as a reflecting mechanism. It monitors the status of each sub-operation, and upon detecting a failure, it utilizes historical interaction data and current observations to diagnose the cause of error and trigger hierarchical recovery strategies. For execution-level errors, it initiates local re-planning, whereas for capability constraints, it escalates to global re-planning to seek collaboration from other robots. This multi-level mechanism effectively resolves errors at the appropriate layer and prevents failures from disrupting the global multi-robot workflow. Together, these agents form a collaborative system that ensures reliable deployment for complex multi-robot manipulation.

Extensive real-world experiments on six tasks validate our framework in three aspects: (1) \textbf{Versatility:} outperforming both learning-based and LLM-based baselines in diverse skills ranging from pick-and-place to articulated object interactions; (2) \textbf{Robustness:} evidenced by a multi-level recovery mechanism that handles physical disturbances via local re-planning and resolves capability constraints through inter-robot collaborative recovery; (3) \textbf{Adaptability:} demonstrated by the seamless transition by our framework from tabletop setups to cross-workspace collaborations involving heterogeneous robots.

\section{Related works}
\subsection{LLMs for Robotics} 
While conventional robotic manipulation models \cite{dp, dp3, pi0, openvla-oft, Mitigating, crosstask, rethinking, cyclemanip} demonstrate promising performance, their understanding of tasks and environments remains constrained. To address this limitation, Large Language Models (LLMs) are increasingly integrated into robotic systems to leverage their strengths in semantic reasoning and high-level planning \cite{LLMzeroshot, LLMfewshot, llmrobot1}. Early works, such as \cite{DoasIcan, PlanningwithWorldModel, InnerMono, llmrobot}, demonstrate the efficacy of LLMs in decomposing semantic instructions into sequential sub-tasks. Leveraging the robust code generation capabilities of LLMs \cite{LLMCode, gptcode, gptcode2}, a specific stream of research focuses on synthesizing executable policies directly. \cite{CaP, ProgPrompt} utilize LLMs to generate Pythonic control logic that invokes predefined action primitives. To improve physical grounding, methods like \cite{Voxposer, Rekep, ReSem3D, omnimanip} formulate manipulation as constraint optimization problems, using the geometric information of the scene, translating natural language into cost maps or spatial constraints to guide motion planning. While these approaches achieve promising manipulation capabilities in single-robot settings, they are inherently limited to individual workspaces. They lack the requisite coordination mechanisms to manage spatio-temporal constraints across multiple robots, making them insufficient for collaborative tasks that demand synchronized execution and long-horizon cooperative behaviors.

In contrast to these single-robot-centric approaches, our framework extends LLMs manipulation paradigm to the multi-robot domain. We introduce a collaborative planner through an agentic workflow that enables heterogeneous robots to achieve parallel execution and cross-workspace collaboration.

\subsection{Multi-Robot Planning and Collaboration}

Prior to the advent of LLMs, multi-robot planning focuses on classical symbolic planning and optimization methods. Traditional multi-robot Task and Motion Planning frameworks typically utilize Planning Domain Description Language (PDDL) or temporal logic to solve long-horizon collaboration problems under geometric constraints \cite{MRTAMP_1, PDDL1, MRpddl1}. While these methods ensure logical correctness and collision-free execution, they rely on rigid, predefined domain rules and struggle to generalize to open-ended semantic instructions. 

Recognizing the limitations of traditional approaches in semantic understanding and adaptability, recent research has extended LLMs capabilities to multi-robot systems \cite{PIP-LLM, lammap, HAMRS, collabot, adhoc, LIP-LLM, Co-navgpt}. A prominent line of work treats multi-robot collaboration as a communication and consensus problem mimicking human teams. For instance, \citet{RoCo}, \citet{DialeticLLM} and \citet{coela} introduce multi-turn dialog mechanisms where robots discuss strategies to reach a consensus on task distribution. Other frameworks explore centralized paradigms. \citet{Smartllm} utilizes a single centralized LLM to decompose instructions into multi-stage task planning. In a more structured approach, \citet{coherent} adopts a centralized hierarchical pipeline that generates sub-task proposals, allocating specific actions to heterogeneous robots based on their capabilities.

Despite their success in high-level planning, most existing LLM-based frameworks limit their validation before execution,  which implicitly assume that the scheduled tasks will be executed successfully. They tend to treat the interaction process as idealized abstract actions, neglecting the execution uncertainties\cite{egotraj, revip} inherent in the physical world. Consequently, these planners lack the requisite dynamic verification and feedback mechanisms during deployment, often leading to cascading system-wide failures when local errors occur.

In contrast to the aforementioned approaches, our work bridges the gap between single-robot manipulation capabilities and multi-robot collaboration. We propose a closed-loop multi-agent architecture that bridges low-level physical manipulation with high-level collaborative task planning. By incorporating a Verification Agent, our framework moves beyond open-loop execution, enabling active outcome monitoring and adaptive failure recovery to ensure robust real-world collaboration.

\begin{figure*}[tbp]
  \centering
  \includegraphics[width=\textwidth]{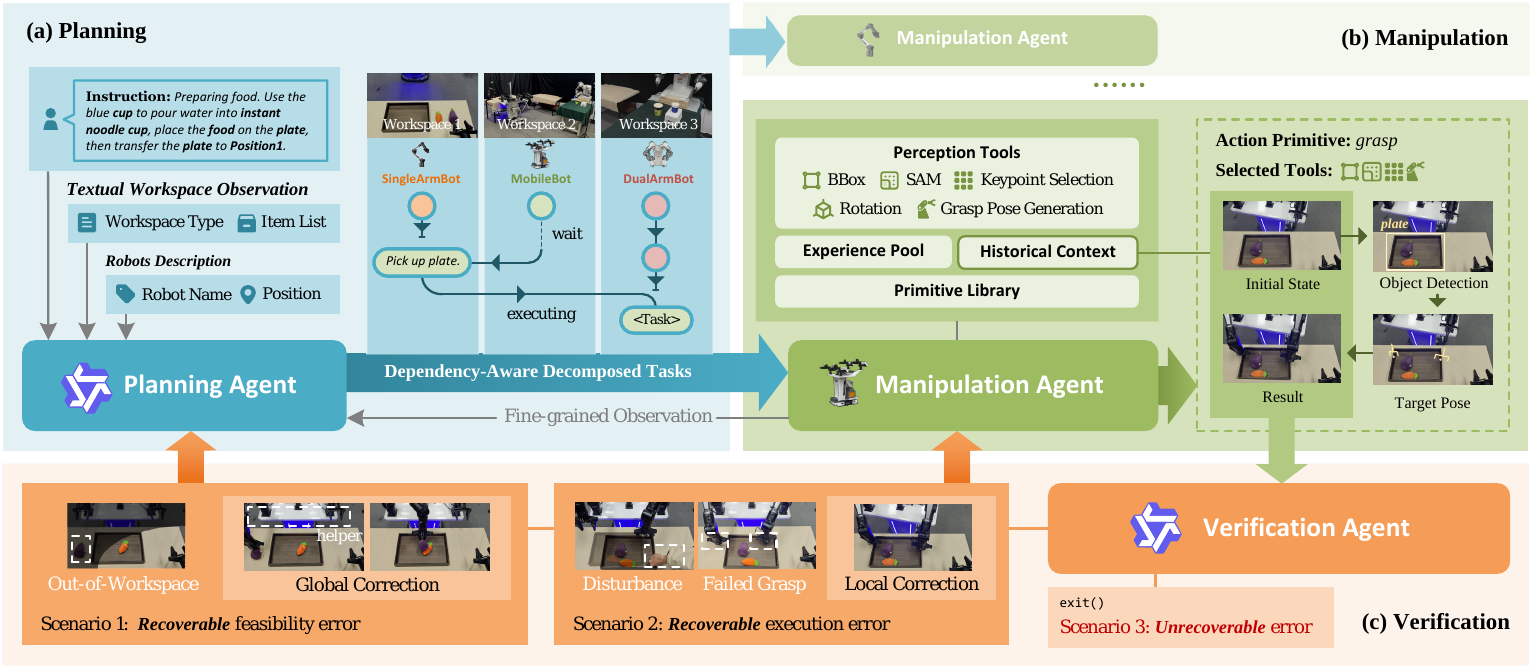}
  \caption{\textbf{Overview of our Closed-Loop Multi-Agent Framework.} The agentic LLM-based workflow integrates three specialized agents: (a) \textbf{The Planning Agent} processes language instructions and coarse-grained workspace observations. It reasons about robot capabilities to generate a dependency-aware task graph, allocating sub-tasks to different robots. (b) \textbf{The Manipulation Agent} grounds these abstract sub-tasks into grounded action primitives using a suite of visual perception tools and historical context. (C) \textbf{The Verification Agent} monitors the physical state to ensure robustness. It implements a hierarchical recovery, handling execution failures through local correction and capabilities violations through global correction. Ensuring robust collaboration in real-world environments.}
  \label{fig:main}
  \vspace{-0.35cm}
\end{figure*}

\section{Method}

\subsection{System Overview}
In this section, we present a closed-loop multi-agent framework designed for reliable multi-robot manipulation in real-world environments. As shown in Fig. \ref{fig:main}, our framework is structured as a collaborative agentic workflow. It takes language instructions and a coarse-grained scene description, which include the robot's position and the general location of objects as input. It integrates semantic planning, physical manipulation, and verification into an iterative cycle coordinated by three specialized agents: the Planning Agent handles high-level reasoning and task decomposition, the Manipulation Agent translates the sub-task into the real world and the Verification Agent assesses action feasibility and execution outcomes. Crucially, these agents dynamically interact to translate abstract instructions into real-world primitives, while robustly adapting to execution-time uncertainties. Detailed designs of each agent and their collaborative mechanisms will be elaborated in the following sections. Specific model selections are detailed in Appendix.\Rmnum{3}.

\subsection{Planning Agent: Semantic Reasoning and Task Allocation}

The Planning Agent serves as the semantic reasoning core of the framework. As shown in Fig. \ref{fig:main}(a), its primary role is to interpret user instructions and generate a high-level task graph that bridges the gap between abstract semantic goals and robot-specific capabilities through two key mechanisms.

\noindent\textbf{Dependency-Aware Task Decomposition.} Given a natural language instruction and coarse-grained scene descriptors, the Planning Agent decomposes the global objective into a \textbf{directed acyclic graph} (DAG) of logical sub-tasks. Unlike generating action sequences, this agent focuses on identifying task dependencies, determining which sub-tasks must be executed sequentially and which can be performed simultaneously. It outputs a schedule of sub-tasks augmented with a \texttt{parallel} flag, allowing robots to perform operations in parallel during this stage which enables concurrent execution by heterogeneous robots to optimize operational efficiency.

\noindent\textbf{Capability-Based Allocation \& Interactive Refinement.} The Planning Agent assigns each sub-task to the most suitable robot by matching the sub-task requirements (e.g., transport) with the heterogeneous robots' capabilities (e.g., mobile base and fixed base). Furthermore, to handle potential uncertainties, the agent is equipped with an interactive refinement capability. In scenarios where the initial instruction or scene information is insufficient (e.g., an ambiguous ``Tidy up'' command or the object to manipulate is unknown), the Planning Agent can conditionally send a \texttt{further\_perception} flag. This triggers a targeted query to the corresponding Manipulation Agent to obtain fine-grained observations, allowing the Planning Agent to refine the task graph based on the updated semantic context.

\subsection{Manipulation Agent: Physical Action Grounding}

The Manipulation Agent functions as the embodied executor within our agentic workflow. As shown in Fig. \ref{fig:main}(b), upon receiving a logical sub-task from the Planning Agent, its objective is to translate this semantic instruction into action primitives that can control the robot directly. To achieve this, the agent leverages a flexible mechanism of adaptive tool invocation. Rather than following a fixed pipeline, it dynamically calls upon a suite of specialized modules, ranging from VLM-based decomposition to visual prompting tools to decompose a sub-operation, perceive targets, and get action parameters for the specific physical context. Specifically, the agent re-perceives the environment between sub-operations in response to potential changes occurring after an action (e.g., opening a drawer).

\noindent\textbf{Role-Based Operation Resolution.} Unlike Planning Agent that decomposes tasks at a high-level, Manipulation Agent focuses on breaking down sub-tasks into executable fine-grained actions. Before execution, the agent first divides sub-task into manipulation steps through VLM-based reasoning. Based on the interaction logic, it not only categorizes entities into active and passive roles but also determines their semantic part descriptions (e.g., ``handle'', ``surface'') for each object. Simultaneously, the agent evaluates object attributes to determine whether a dual-arm action is required at this stage. This structured resolution ensures that subsequent tools focus on the correct entities.

\noindent\textbf{Tool-Augmented Visual Perception.} To bridge the gap between semantic roles and action parameters, the agent executes a comprehensive visual grounding pipeline utilizing the part descriptions extracted in the previous stage. \textit{Object Detection \& Segmentation:} The agent first invokes a VLM-based detector to identify bounding boxes for active and passive objects, followed by the Segment Anything Model (SAM) \cite{SAM} to generate precise pixel-level masks. \textit{Affordance-Aware Keypoint Selection:} The agent employs a visual prompting tool \cite{ReSem3D} to localize optimal interaction points, such as the upper part of a cup handle. For dual-arm grasping, the agent applies a geometric heuristic to identify the key points for center symmetry of objects to ensure stability. Crucially, these 2D keypoints are projected into 3D space using depth information, establishing the spatial anchors required for manipulation.

\noindent\textbf{Action Primitives Generation with Adaptive Tool Invocation.} Once the spatial keypoints are established, the agent synthesizes the operation by querying an action primitive library containing skills in Tab \ref{tab:primitives}. During this generation process, the agent adaptively invokes specialized tools if a primitive requires physical grounding. For instance, when instantiating a grasp primitive, semantic keypoints alone may be insufficient for physical stability. Current 6-DoF grasp generation methods are diverse \cite{taskgrasp, igrasp, anygrasp, omnidexgrasp}, the agent triggers AnyGrasp \cite{anygrasp} to generate candidate 6-DoF grasping poses (details are provided in Appendix.\Rmnum{2}.C), which are then matched against the extracted 3D semantic keypoints to select the optimal pose that minimizes spatial alignment error. In addition to spatial precision provided by the keypoints selection, many manipulation tasks require appropriate object orientation to satisfy specific semantic requirements. To addressing this, the agent can activate the rotation perception tool. Leveraging rigid-body assumption, this tool projects the robot's end-effector frame onto the object's center and a VLM then performs spatial reasoning to analyze the object's current state and the task requirements, predicting the necessary rotation axis, direction, and angle. This allows the agent to execute intuitive reorientation actions.

\begin{table}[tbp]
  \centering
  \caption{\textbf{Library of Action Primitives.} Modes indicate support for Single-arm (S) or Dual-arm (D) execution.}
  \label{tab:primitives}
  \resizebox{\linewidth}{!}{
  \begin{tabular}{@{}l c l@{}}
    \toprule
    \textbf{Primitive} & \textbf{Mode} & \textbf{Function} \\
    \midrule
    \multicolumn{3}{@{}l}{\textit{\textbf{Basic Manipulation}}} \\
    \texttt{Grasp} & S / D & Execute a grasp and lift up the object. \\
    \texttt{Place} & S / D & Move to the target, put down and release gripper. \\
    \texttt{Lift Up} & S / D & Vertical motion ($+\Delta z$). \\
    \texttt{Put Down} & S / D & Vertical motion ($-\Delta z$). \\
    \texttt{Reset Home} & S / D & Retreat to a predefined safe configuration. \\
    \midrule
    \multicolumn{3}{@{}l}{\textit{\textbf{Spatial Adjustment}}} \\
    \texttt{Move XY} & S / D & Planar translation while maintaining height. \\
    \texttt{Move Pose} & S & Move to $(x,y,z)$ keeping current pose. \\
    \texttt{Rotate} & S & Re-orient the object around the predicted axis. \\
    \texttt{Align} & S & Minimize the distance between semantic keypoints. \\
    \midrule
    \multicolumn{3}{@{}l}{\textit{\textbf{Constraint Interaction}}} \\
    \texttt{Open} & S & Follow the articulation trajectory (prismatic). \\
    \texttt{Close} & S & Restore the articulated object to a closed state. \\
    \bottomrule
  \end{tabular}
  }
  \vspace{-0.5cm}
\end{table}

\noindent\textbf{Memory-Augmented Reasoning.} To enhance the manipulation efficiency and reduce the ambiguity in sub-tasks description, we implemented a dual-layer memory mechanism. First, for the short-term memory, the agent maintains an interaction history of previous manipulation steps. It will be used in the sub-task decomposition phase which resolves ambiguous sub-task instructions and verification phase. For example, when receiving ``Pour water", it can determine from historical manipulations whether the grasping action has already been performed or if only the pouring action needs to be executed. Second, for manipulation efficiency, we implement an experience pool for the sub-task decomposition and action primitives generation. For a sub-task, the agent first identifies the actions required for the task and the objects involved in the interaction and records them as a template signature (e.g., \texttt{<Action> Active $\to$ Passive}). After this sub-task is successfully executed, the agent will record the results of sub-task decomposition and the corresponding action primitives. Next time the agent encounters a sub-task exactly matching the signature, it triggers a shortcut, using the validated sub-task decomposition and the primitives directly. The experience pool reduces the number of VLM API calls which accelerates execution time, enabling manipulation to proceed directly into the perception phase.

\subsection{Verification Agent: Local Monitoring and Hierarchical Recovery}

The Verification Agent acts as the reflection component in the agentic workflow to ensure system reliability. As shown in Fig. \ref{fig:main}(c), it transforms the linear execution into a robust closed-loop system by monitoring the task states and leveraging a hierarchical recovery mechanism. Each Manipulation Agent is equipped with a Verification Agent which obtains the execution status and triggers different strategies based on the types of failures: executing a local self-correction when execution fails, while interacting with the Planning Agent for global re-planning when encountering feasibility errors.

\noindent\textbf{Discrete Visual Outcome Validation.} To verify each sub-operation acted by the Manipulation Agent, Verification Agent employs a visual feedback tool to perform a discrete and semantic assessment. Upon the completion of a sub-operation, the agent analyzes pre-action and post-action images relative to the instructions, performing logic validation to generate a binary success prediction.

\noindent\textbf{Hierarchical Error Recovery.} When a failure is detected, the Verification Agent diagnoses the error type and triggers recovery at the appropriate system level. For execution failures, where the task remains feasible (e.g., a grasping slip or misalignment), the agent initiates a local recovery loop without disturbing the global plan. By referencing the interaction history and the visual states, the agent autonomously constructs a corrective sequence to re-establish the necessary states. These recovery steps are dispatched directly to the Manipulation Agent, forming an efficient inner loop for rapid self-correction. 

Conversely, if an action is identified as physically infeasible, typically due to workspace constraints or missing objects , the Verification Agent will send the failure to the Planning Agent, triggering a global recovery loop. It will submit a failure feedback containing the specific error context and state, which activates re-allocation rules in  the Planning Agent. Consequently, the Planning Agent generates a new task graph based on the cause of the error and the current state of robots, while continuing to pursue the original task objectives. For example, when a fixed-based robot arm can't pick up an object out of its working space, Planning Agent will dispatch a mobile robot to solve the problem. This hierarchical mechanism ensures task continuity through agentic collaboration, preventing single-robot failures from stopping the entire workflow.

\section{Experiments}

\begin{figure*}[htbp]
    \centering
    \includegraphics[width=0.95\textwidth]{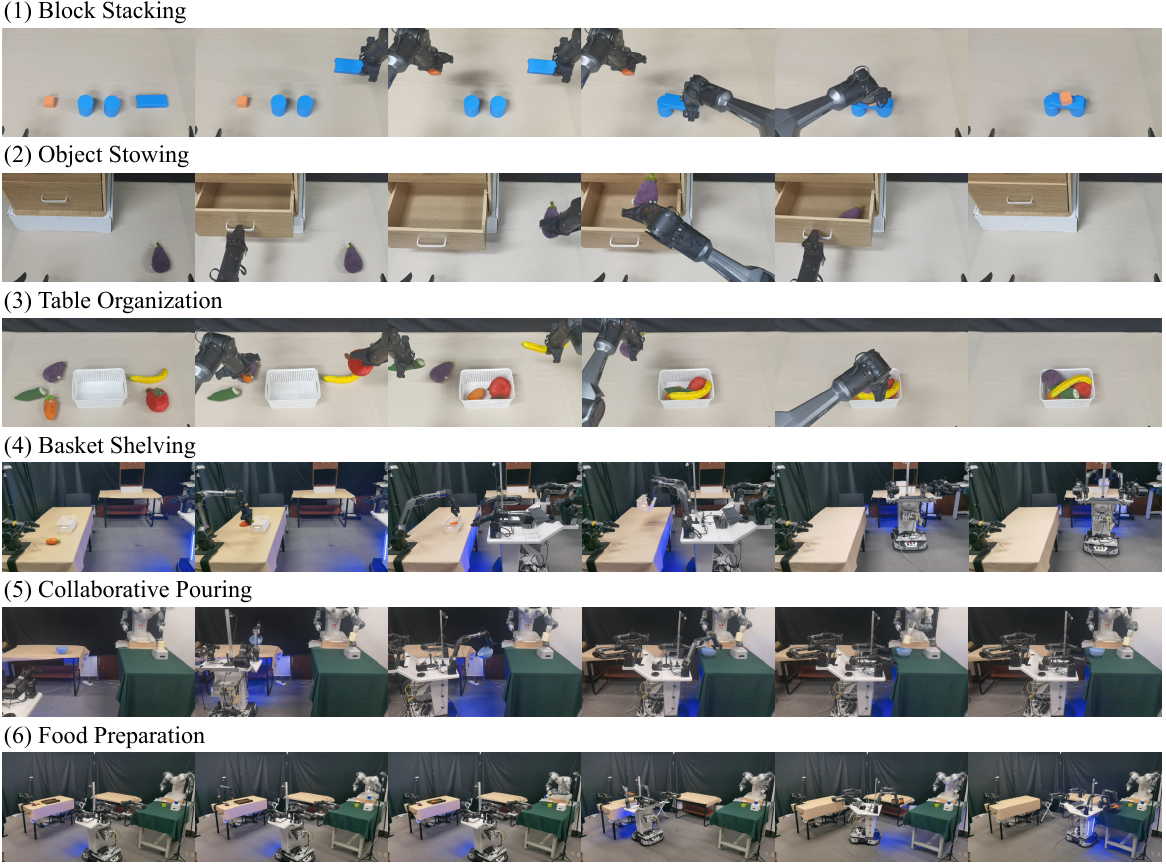}
    \caption{\textbf{Execution Process of Six Real-world Experiments.} We validate our framework on (1-3) dual-arm manipulation tasks and (4-6) multi-robot collaborative scenarios, demonstrating the framework's versatility across varying complexities and robot configurations.}
    \label{fig:exp}
    \vspace{-0.35cm}
\end{figure*}

In this section, we evaluate the effectiveness of our proposed framework through extensive real-world experiments. We designed 6 distinct tasks of varying complexity, ranging from single-robot manipulation to heterogeneous multi-robot collaboration. 
The task execution processes are shown in Fig.\ref{fig:exp}.

\subsection{Experimental Setup}

\textbf{Hardware Configuration:} Our system comprises three types of robots: (1) an Agilex Cobot Magic dual-arm mobile robot, (2) an ABB-YuMi dual-arm desktop robot, and (3) an Agilex Piper single-arm desktop robot. Based on the number of robots, we have categorized 6 tasks into two categories:
\paragraph{Table Top Dual-Arm Tasks} (1) Block Stacking: A precision task requiring the robot to place a rectangular block across two cylinders (bridge) and stack a cube on the top. (2) Table Organization: The robot needs to tidy up a workspace by multi-round pick \& place 5 odds and ends into a basket sequentially under fuzzy instructions without knowing the objects on the table. (3) Object Stowing: An articulated object manipulation task which contains multiple stages involving opening a drawer, placing an object inside, and closing drawer.
\paragraph{Multi-Robot Collaboration} (1) Basket Shelving:  A Piper robot places an object into a basket. The Cobot Magic robot lifts the basket with dual-arm and transports it to a shelf in a different workspace. (2) Collaborative Pouring: Receiving an instruction to pour a bowl of water, Cobot Magic first needs to move to a table which has a bowl on it, transport it to ABB's table, then place it in an appropriate place for ABB to pour water into it. (3) Food Preparation: A three-robot task. The ABB robot pours water into an instant noodle cup and places the cup back. Simultaneously, the Piper places two food objects on the plate on a different table. Finally, the Cobot Magic transports the prepared food plate from Piper's table to the ABB's table.

\subsection{Dual-arm Table Top Task}
We validate our framework's manipulation ability against (1)  OpenVLA-OFT \cite{openvla-oft} and (2) $\pi_{0}$ \cite{pi0}, representing mainstream learning-based end-to-end policies. (3) ReKep \cite{Rekep}, a learning-free method that employs VLMs to optimize relational keypoint constraints for robot manipulation. We report success trials and task completion (Comp. \%) over 20 trials for each task.

\noindent\textbf{Experimental Setup.} Both learning-based baselines (OpenVLA-OFT and $\pi_{0}$) are fine-tuned in a single-task setting using 50 demonstrations per task. To isolate perception errors and strictly evaluate manipulation capabilities, we provide ReKep with ground-truth keypoint annotations manually during execution. 

\begin{table}[t]
\centering
\caption{\textbf{Table Top Tasks Performance Comparison.} 
We evaluate OpenVLA-OFT, $\pi_{0}$ (learning-based), ReKep (LLM-based), and our method (Ours) across three tasks. For each method and task, we report: (1) number of successful trials out of 20, and (2) average task completion percentage.}
\label{tab:per_task_comparison_vertical}

\setlength{\tabcolsep}{2pt}

\begin{tabular}{l c c c c c c}
\toprule
\multirow{2}{*}{\textbf{Method}} & 
\multicolumn{2}{c}{\textbf{Block Stacking}} & 
\multicolumn{2}{c}{\textbf{Table Org.}} & 
\multicolumn{2}{c}{\textbf{Object Stowing}} \\
\cmidrule(lr){2-3} \cmidrule(lr){4-5} \cmidrule(lr){6-7}
& Succ. & Comp.(\%) & Succ. & Comp.(\%) & Succ. & Comp.(\%) \\
\midrule

OpenVLA-OFT \cite{openvla-oft} & 1/20 & 35 & 5/20 & 55 & 4/20 & 31 \\

$\pi_{0}$ \cite{pi0}               & 9/20 & 46 & 9/20 & 75 & 8/20 & 60 \\
\addlinespace[2pt]

ReKep Annot. \cite{Rekep}       & 7/20 &  56 & 6/20 & 44 & 0/20 & 0 \\
\addlinespace[2pt]

Ours               & \textbf{14/20} & \textbf{85} & \textbf{11/20} & \textbf{76} & \textbf{14/20} & \textbf{78} \\

\bottomrule
\end{tabular}
\vspace{-0.5cm}
\end{table}

As summarized in Tab. \ref{tab:per_task_comparison_vertical}, our method significantly outperforms all baselines on three tasks. The learning-based methods struggled in contact-rich tasks like \textit{Block Stacking} and \textit{Object Stowing}. Lacking precise geometric grounding, minor deviations in the predicted end-effector's pose frequently led to hard collisions with blocks and drawer, causing immediate failures. In long-horizon task \textit{Table Organization}, baselines often fail to grasp the target object yet continue to execute the subsequent trajectory, letting these execution errors remain undetected and unchecked throughout the sequence. 
The learning-free method ReKep completely failed in \textit{Object Stowing} because their original method could not define the end key points when operating drawers. In \textit{Block Stacking}, the optimization solver often returns sub-optimal solutions that failed to strictly avoid collisions while attempting to reach the goal. 
In contrast, our framework achieved the highest success rates across all tasks, demonstrating that our tool-augmented perception ensures the precision required for stacking, the diverse combination action primitives provide the necessary versatility, and the Verification Agent provides the closed-loop robustness necessary to detect and try to recover from the execution errors.

\begin{table}[t]
\centering
\caption{\textbf{Table Top Tasks Performance Comparison under External Disturbances.} Comparison of Success Rates (SR) across three desktop tasks. Baselines include OpenVLA-OFT and $\pi_{0}$.}
\label{tab:desktop_robustness}

\setlength{\tabcolsep}{5.5pt}

\begin{tabular}{l c c c c}
\toprule
\textbf{Method} & \textbf{Stacking} & \textbf{Table Org.} & \textbf{Stowing} & \textbf{Avg. (\%)} \\
\midrule
OpenVLA-OFT & 0/10 & 0/10 & 2/10 & 7 \\
\addlinespace[2pt]
$\pi_{0}$        & 2/10 & 1/10 & 3/10 & 20 \\
\addlinespace[2pt]
\midrule
Ours & \textbf{5/10} & \textbf{6/10} & \textbf{8/10} & \textbf{63}\\
\bottomrule
\end{tabular}
\vspace{-0.1cm}
\end{table}

\subsection{Robustness to External Disturbances}
To evaluate the robustness of our framework against real-world uncertainties, we designed three distinct disturbances ranging from local interference to global failure. We bring these disturbances at random intervals during the execution of both desktop and multi-robot tasks. 
\paragraph{Action Level} Human operators remove or displace the target object while the robot is attempting to interact, simulating environmental changes or interaction failure.
\paragraph{Sub-task Level} After a sub-task is completed, the operator resets the environment state, specifically testing its ability to execute multi-step recovery sequences. 
\paragraph{Global Level} In multi-robot task, a failure may occur if the target is placed outside the workspace of a fixed-base robot. The system must invoke a collaborative recovery and dispatch other robots to assist in the manipulation.

As shown in Tab. \ref{tab:desktop_robustness}, learning-based baselines struggled significantly under these perturbations, resulting in low average success rates. Due to the absence of a closed-loop feedback mechanism, these policies often continued to execute subsequent trajectories despite environmental change. While they may attempt a simple re-grasp if a failure occurs during the initial phase, they fail to recover from complex disturbances that require multi-step reasoning. In contrast, our method maintained a robust average success rate of \textbf{63\%}. Compared to the undisturbed condition, the success rate shows only a slight decrease, demonstrating the critical necessity of our verification loop.

\begin{table}[t]
\centering
\caption{\textbf{System Robust Analysis under External Disturbances.} We evaluate the closed-loop pipeline under external disturbances. Metrics include Failure Identification rate (ID), Re-planning success rate (Re-Plan), Physical Recovery success rate (Phy-Rec), and the Final Task Success Rate (SR).}
\label{tab:full_recovery_analysis}
\setlength{\tabcolsep}{4pt}

\begin{tabular}{l l c c c c}
\toprule
\multirow{2}{*}{\textbf{Task}}  & \multirow{2}{*}{\textbf{Scenario}} &\textbf{ID} & \textbf{Re-Plan} & \textbf{Phy-Rec} & \textbf{Final SR} \\
 &  & (\%) & (\%) & (\%) & (\%) \\
\midrule
\multirow{3}{*}{\textit{Dual-Arm}} & Block Stacking & 90 & 100 & 78 & 50 \\
\addlinespace[2pt]
& Table Organization & 100 & 100 & 80 & 60 \\
\addlinespace[2pt]
& Object Stowing & 90 & 100 & 100 & 80 \\
\midrule
\textit{Multi-Robot}   &  Basket Shelving  & 90 & 100 & 89 & 40 \\
\bottomrule
\end{tabular}

\vspace{4pt}
\vspace{-0.5cm}
\end{table}

\subsection{Verification Agent Analysis}
To demonstrate the reliability of our Verification Agent, we tracked four metrics under the aforementioned disturbance conditions: Failure Identification Rate (ID), Re-planning Success Rate (Re-Plan), Physical Recovery Success Rate (Phy-Rec) which quantifies the robot's ability to physically resolve the disturbance conditioned on a valid plan and Final Task Success Rate (Task SR). The results are shown in Tab. \ref{tab:full_recovery_analysis}. The Verification Agent shows robust failure identification capabilities around \textbf{90--100\%} failure identification rate across all tasks. This confirms that our VLM-based verification module effectively distinguishes between successful outcomes and varying types of errors. 

In single-robot tasks, the system primarily relies on an internal recovery loop. The Re-planning Success Rate was consistently \textbf{100\%}. However, there is a gap between detection and final success in execution, as irreversible states may occur during task execution, such as collapse of the building blocks, which require more meticulous operational skills to recover. In contrast, \textit{Object Stowing} achieved a high 80\% Final SR, proving the system's effectiveness in handling reversible state regressions. In the multi-robot task, we validate the global recovery capability of our framework in the \textit{Basket Shelving} task. We placed an object outside the fixed-base robot's workspace, making it unreachable for the required operation. The system successfully identified the error caused by exceeding the workspace and performed global re-planning with \textbf{100\%} re-planning success. The only exception occurred when the object moved out of the field of view after a failed placement operation.

\begin{figure}[t]
    \centering
    \includegraphics[width=\columnwidth]{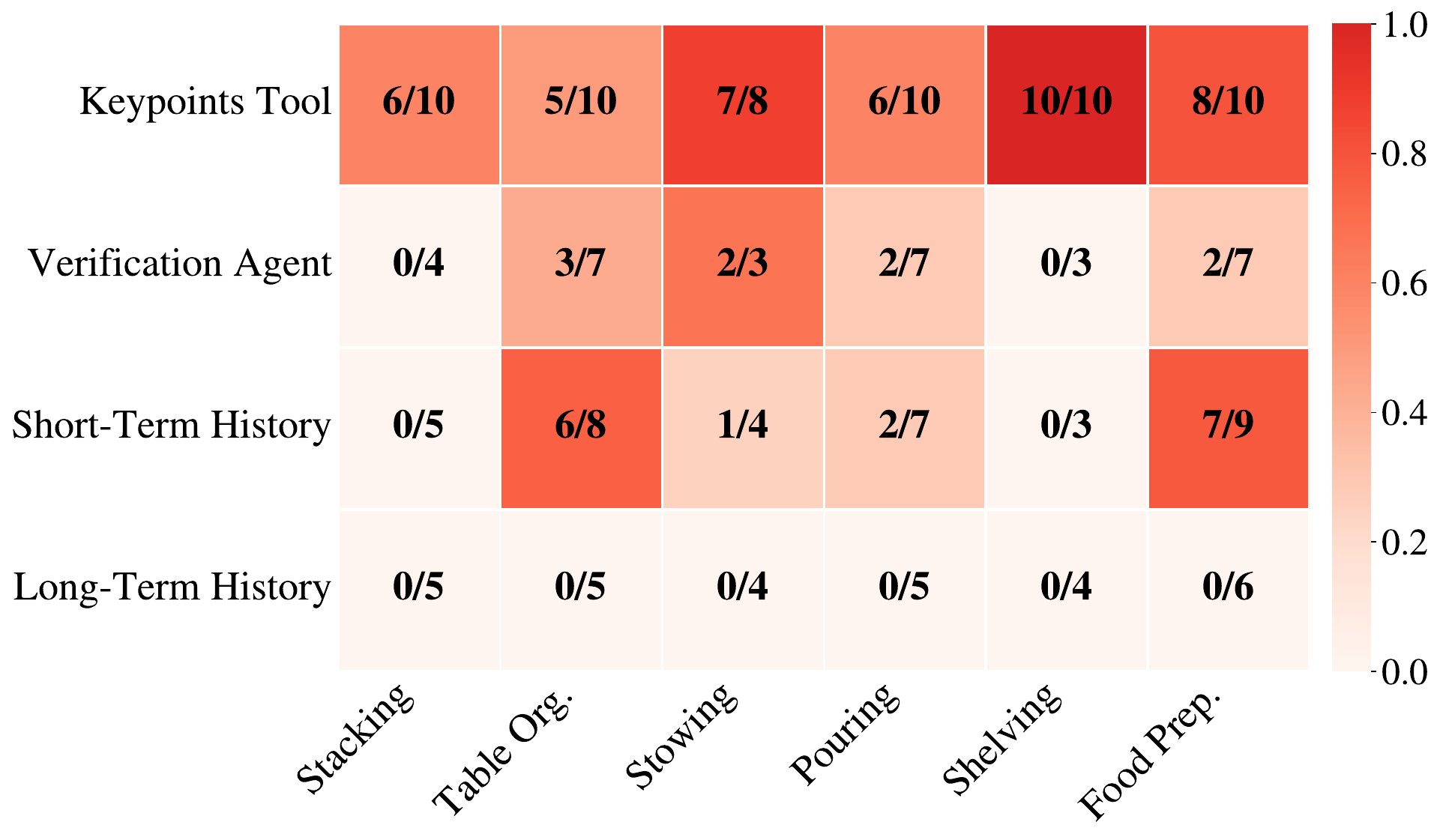}
    \caption{\textbf{Analysis of Failure Contributions by Module Ablation.} We visualize the impact of removing specific components on task failures. The ratio ($x/y$) denotes the number of failures ($x$) directly attributed to the ablated module relative to the total number of failures ($y$) observed in that task.}
    \label{fig:ablation}
    \vspace{-0.5cm}
\end{figure}

\begin{table*}[t]
\centering
\caption{\textbf{Ablation Study on Key Components.} 
We evaluate the impact of: (1) Keypoint Perception Tool, (2) Verification Agent, (3) Short-Term Memory, and (4) Long-Term Memory. The full model includes all components. Metrics: Success Rate (SR) and Task Completion (Comp. \%).}
\label{tab:ablation}

\setlength{\tabcolsep}{2.5pt}

\begin{tabular}{l c c c c c c c c c c c c}
\toprule
\multirow{2}{*}{\textbf{Model Variant}} & 
\multicolumn{2}{c}{\textbf{Block Stacking}} & 
\multicolumn{2}{c}{\textbf{Table Org.}} & 
\multicolumn{2}{c}{\textbf{Object Stowing}} & 
\multicolumn{2}{c}{\textbf{Collab. Pouring}} & 
\multicolumn{2}{c}{\textbf{Basket Shelving}} & 
\multicolumn{2}{c}{\textbf{Food Prep.}} \\
\cmidrule(lr){2-3} \cmidrule(lr){4-5} \cmidrule(lr){6-7} \cmidrule(lr){8-9} \cmidrule(lr){10-11} \cmidrule(lr){12-13}
& Succ. & Comp. (\%) & Succ. & Comp. (\%) & Succ. & Comp. (\%) & Succ. & Comp. (\%) & Succ. & Comp. (\%) & Succ. & Comp. (\%) \\
\midrule

\textbf{Full Model (Ours)}                     & 14/20 & 85 & 11/20 & 76 & 14/20 & 78 & 11/20 & 79 & 12/20 & 84 & 10/20 & 69 \\
\addlinespace[2pt]
\midrule[0.1pt]
\addlinespace[2pt]

-- w/o Keypoint Perception Tool       & 0/10 & 35 & 0/10 & 32 & 2/10 & 45 & 0/10 & 16 & 0/10 & 50 & 0/10 & 26  \\
\addlinespace[2pt]

-- w/o Verification Agent             & 6/10 & 85 & 3/10 & 71 & 7/10 & 80 & 3/10 & 70 & 7/10 & 87 & 3/10 & 56 \\
\addlinespace[2pt]

-- w/o History in Planning            & 5/10 & 78 & 2/10 & 46 & 6/10 & 83 & 3/10 & 61 & 7/10 & 95 & 1/10 & 51 \\
\addlinespace[2pt]

-- w/o Exp.                           & 5/10 & 78 & 5/10 & 65 & 6/10 & 80 & 5/10 & 84 & 6/10 & 80 & 4/10 & 70 \\

\bottomrule
\end{tabular}
\vspace{-0.25cm}
\end{table*}

\subsection{Ablation Study}
To analyze the contributions of each module within our framework, we conducted an ablation study across 6 tasks. We evaluated the system by systematically removing: (1) the visual perception tool, (2) the Verification Agent, (3) short-term memory (interaction history recorded by the Manipulation Agent), and (4) long-term memory (experience pool). Quantitative results are summarized in Tab. \ref{tab:ablation}, and Fig. \ref{fig:ablation} details module contributions to task failures.

\noindent\textbf{Impact of visual perception tool.} We employ the same visual perception method from ReKep\cite{Rekep} to replace our process, which uses position and feature clustering to obtain representative key points throughout the image. As shown in Fig. \ref{fig:ablation}, the absence of the object-specific keypoint selection process resulted in a catastrophic performance drop. The results reveal the problems of previous multi-robots planning methods, they treat execution as idealized process, ignoring that physical manipulation is coupled with actual perceptual process. For LLM-based manipulation approaches, the precise keypoint selection serves as the essential bridge connecting the high-level plan to the real-world environment. It is not merely an auxiliary step but essential for any successful manipulation.

\noindent\textbf{Impact of Verification Agent.} The ablation of the Verification Agent led to a decline particularly in long-horizon tasks like \textit{Food Preparation} and \textit{Table Organization}. In these tasks, every individual operation carries a non-zero failure probability. Without Verification Agent, these minor execution errors accumulate, eventually becoming fatal to the overall task. This validates that a closed-loop mechanism for error detection and recovery after each execution is essential for the successful completion of complex multi-robot tasks.

\noindent\textbf{Impact of Short-Term Memory.} Removing short-term memory caused a drop in sequential manipulation tasks, most notably in \textit{Table Organization} and \textit{Food Preparation}. To address changes in the environment following an operation, we decompose the sub-operation into action primitives. Therefore, we observed that without historical context, some sub-task instructions might be ambiguous, such as ``place'' command could mean that an object can be placed directly, or needs to be grasped before being placed. This was particularly evident in pouring sequences. Short-term memory eliminates these redundancies by recording the interaction history and state, helping Manipulation Agent generate sub-operations that are logically consistent with the robot's current status.

\noindent\textbf{Impact of Long-Term Memory.} For the long-term memory module, our primary objective is to enhance manipulation efficiency and reduce the frequency of LLM invocations. Experimental results indicate that removing this module only results in a slight decline and contributes to zero failures in method performance. This demonstrates the robustness of our Manipulation Agent in decomposing sub-tasks and generating action primitives, proving the framework does not strictly rely on memorization. Second, it demonstrates that the design of the experience pool enhances stability. By retrieving validated actions for repetitive sub-tasks, the system avoids the variance introduced by VLM generation, ensuring higher consistency and efficiency in execution.

\begin{figure}[htbp]
    \centering
    \includegraphics[width=0.90\columnwidth]{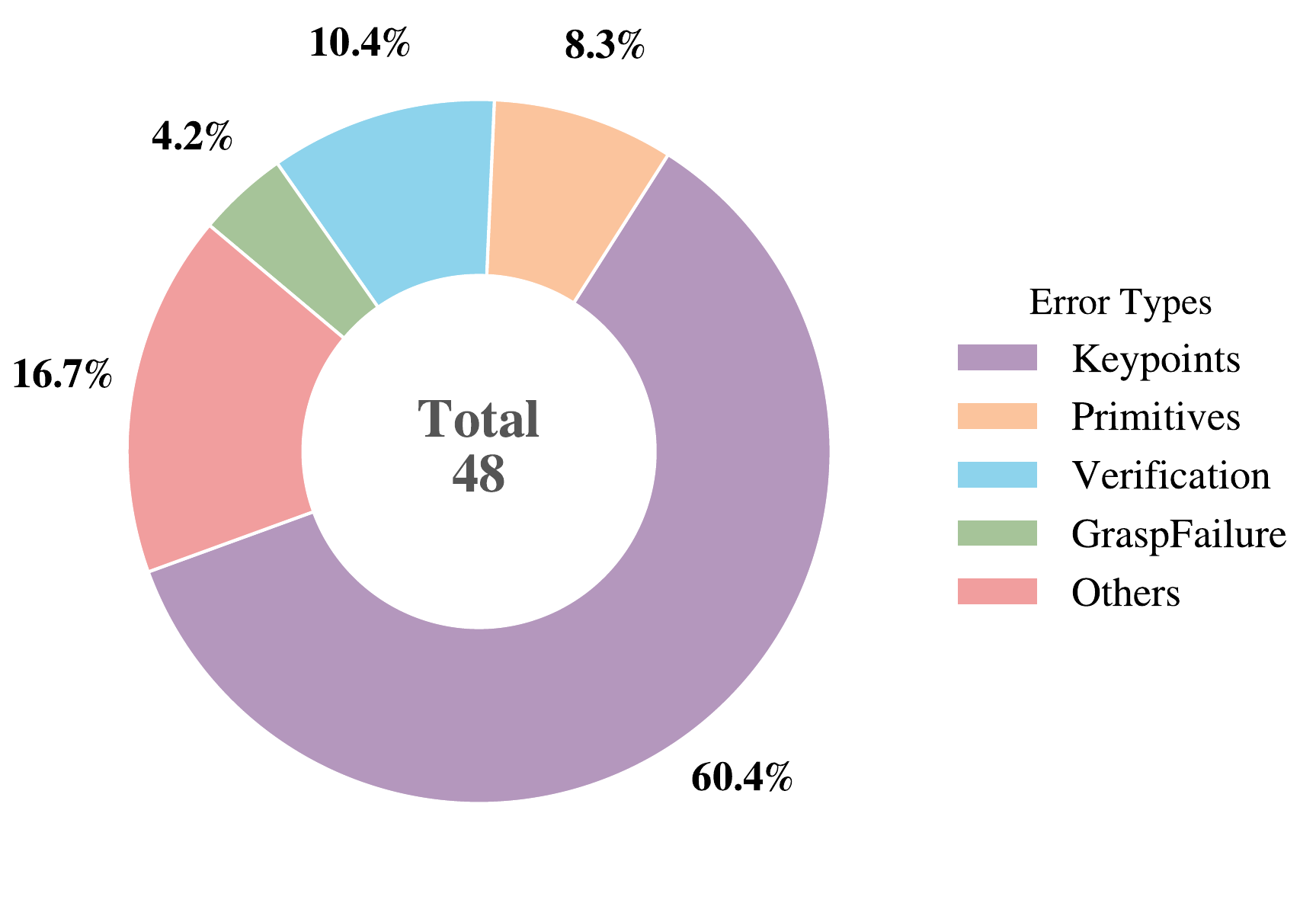}
    \caption{\textbf{Error Breakdown in System Deployment.} We analyzed the total 48 failure cases encountered by the Full Model across all tasks reported in Tab. \ref{tab:ablation}.}
    \label{fig:error}
    \vspace{-0.5cm}
\end{figure}

\subsection{System Error Breakdown}

To identify the limitations of our current framework, we record the failure distribution across individual modules based on the results from the full model experiments in Tab. \ref{tab:ablation}. As illustrated in Fig. \ref{fig:error}, keypoint selection constitutes the dominant failure mode. This bottleneck arises because the VLM struggles to localize keypoints that precisely satisfy the semantic constraints, resulting in a mismatch between the VLM's reasoning and the actual environment. Action primitives generation failures followed, sometimes the robot failed to find a kinematic solution or the VLM failed to correctly reason the relation between keypoints, resulting in incorrect parameter input. Beyond these two modules, failures such as Verification Agent errors and grasp failures account for a low percentage. Additionally, irrecoverable failures occasionally arise due to physical constraints, such as robotic arm collisions, noisy sensor data from depth cameras, and occasional navigation drift. However, each of these failures represents a relatively minor portion of the total cases.

\section{Conclusions}

In this paper, we propose a closed-loop multi-agent framework to enable robust and collaborative manipulation for a multi-robot system. By structuring the system into three specialized agents-Planner, Manipulation, and Verification, we bridge the gap between high-level reasoning and low-level physical execution. Extensive experiments across fine-grained desktop tasks to multi-robot collaborative tasks validate our framework possesses high versatility in executing diverse skills, ranging from precision pick-and-place to articulated object interactions. Furthermore, our framework demonstrates robustness by maintaining continuous multi-robot workflows under disturbances through hierarchical recovery. Finally, transition from desktop setups to cross-workspace collaborations confirms the framework's adaptability in scaling from single-robot to heterogeneous multi-robot systems.

\noindent\textbf{Limitations and Future Work.} Currently, our system relies on a library of action primitives, which limits its ability to handle more complex manipulation tasks. Additionally, the current workflow is constrained by the execution speeds of heterogeneous robots. The system primarily employs a dependency check, lacking a dynamic scheduling mechanism to optimize the global efficiency. In future work, we aim to integrate learning-based primitive discovery and explore concurrent task allocation methods to minimize idle time and enhance the efficiency of multi-robot collaboration.

\section*{Acknowledgments}
This work was supported partially by NSFC (92470202), Guangdong NSF Project (No.2023B1515040025), Guangdong Key Research and Development Program (No.\-2024B0101040004, No.2025B0909020002).

\bibliographystyle{plainnat}
\bibliography{references}
\clearpage
\appendix
\input{appendix}

\end{document}

%% file: appendix.tex
\section{Hardware Setup}
\begin{figure*}[tbp]
  \centering
  \includegraphics[width=\textwidth]{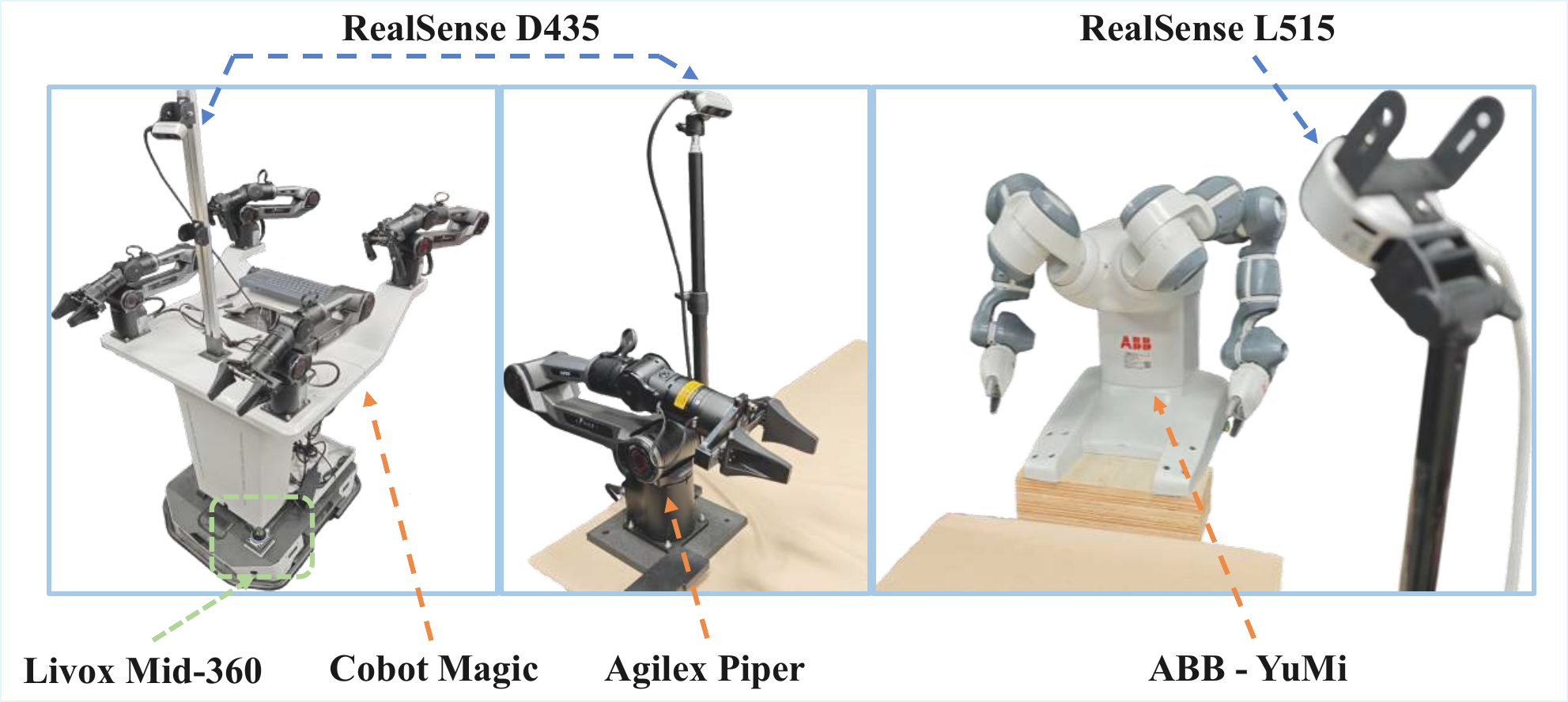}
  \caption{\textbf{Overview of our multi-robot platform.} We validate our framework across three heterogeneous hardware configurations: 
  (a) The \textit{Mobile Dual-Arm Platform (Cobot Magic)}, featuring two 6-DoF arms mounted on a two-wheeled differential chassis with an onboard Intel RealSense D435 camera. 
  (b) The \textit{Single-Arm Platform (Agilex Piper)}, a lightweight 6-DoF manipulator with a third-person Intel RealSense D435 camera. 
  (c) The \textit{Dual-Arm Platform (ABB YuMi)}, a high-precision IRB 14000 robot paired with an Intel RealSense L515 camera.}
  \label{fig:devices}
  \vspace{-1.7cm}
\end{figure*}

We employ three distinct robotic platforms to validate our framework (shown in Fig. \ref{fig:devices}). To ensure a unified action space, all action primitives operate via the End-Effector (EEF) pose commands. These pose targets are translated into joint angles using inverse kinematics (IK) solvers for execution. The detailed configurations are as follows:

\paragraph{\textbf{Mobile Dual-Arm Platform (Agilex Cobot Magic)}}
This platform features two 6-DoF arms on a two-wheeled differential chassis equipped with a Livox Mid-360 LiDAR. Leveraging the LiDAR data, we employ the FAST-LIO\cite{fast_lio} algorithm for mapping and localization, while the ROS1 navigation stack is utilized for global path planning and navigation. Visual perception is provided by an Intel RealSense D435 camera. For low-level control, we utilize the PyRoki\cite{kim2025pyrokimodulartoolkitrobot} library for inverse kinematics solving. 

\paragraph{\textbf{Single-Arm Platform (Agilex Piper)}}
The Agilex Piper is a 6-DoF arm with a 1.5 kg payload. Visual perception is provided by an Intel RealSense D435 camera. Similar to the mobile platform, its IK solving is handled by PyRoki\cite{kim2025pyrokimodulartoolkitrobot}.

\paragraph{\textbf{Dual-Arm Platform (ABB YuMi)}}
We use the ABB YuMi (IRB 14000) for high-precision tasks, with an overhead Intel RealSense L515 camera. The robot is interfaced using YumiPy, which handles communication and IK solving.

\section{Perception Tools Details}
\label{appendix:perception_tools}

\subsection{VLM-based Object Detection and Segmentation}
To achieve precise object manipulation, our system employs a coarse-to-fine perception strategy. We first utilize a Vision-Language Model (VLM) to obtain initial bounding boxes, which subsequently serve as spatial prompts for the Segment Anything Model (SAM) to generate pixel-wise masks. The object detection module bridges the high-level task planner and the low-level segmentation module through two key stages. 

First, in the \textit{Detection via VLM} stage, we employ Qwen3-VL-Plus to detect the identified targets within the current RGB observation. The model is queried with a structured instruction to identify the specified objects and return their 2D bounding boxes (BBoxes) in JSON format. 
Subsequently, during \textit{Segmentation}, the system performs a post-processing step to associate the detected bounding boxes with their semantic roles. These structured BBoxes and their associated roles are then passed to the SAM module as prompt inputs for fine-grained segmentation.

\begin{figure*}[tbp]
  \centering
  \includegraphics[width=\textwidth]{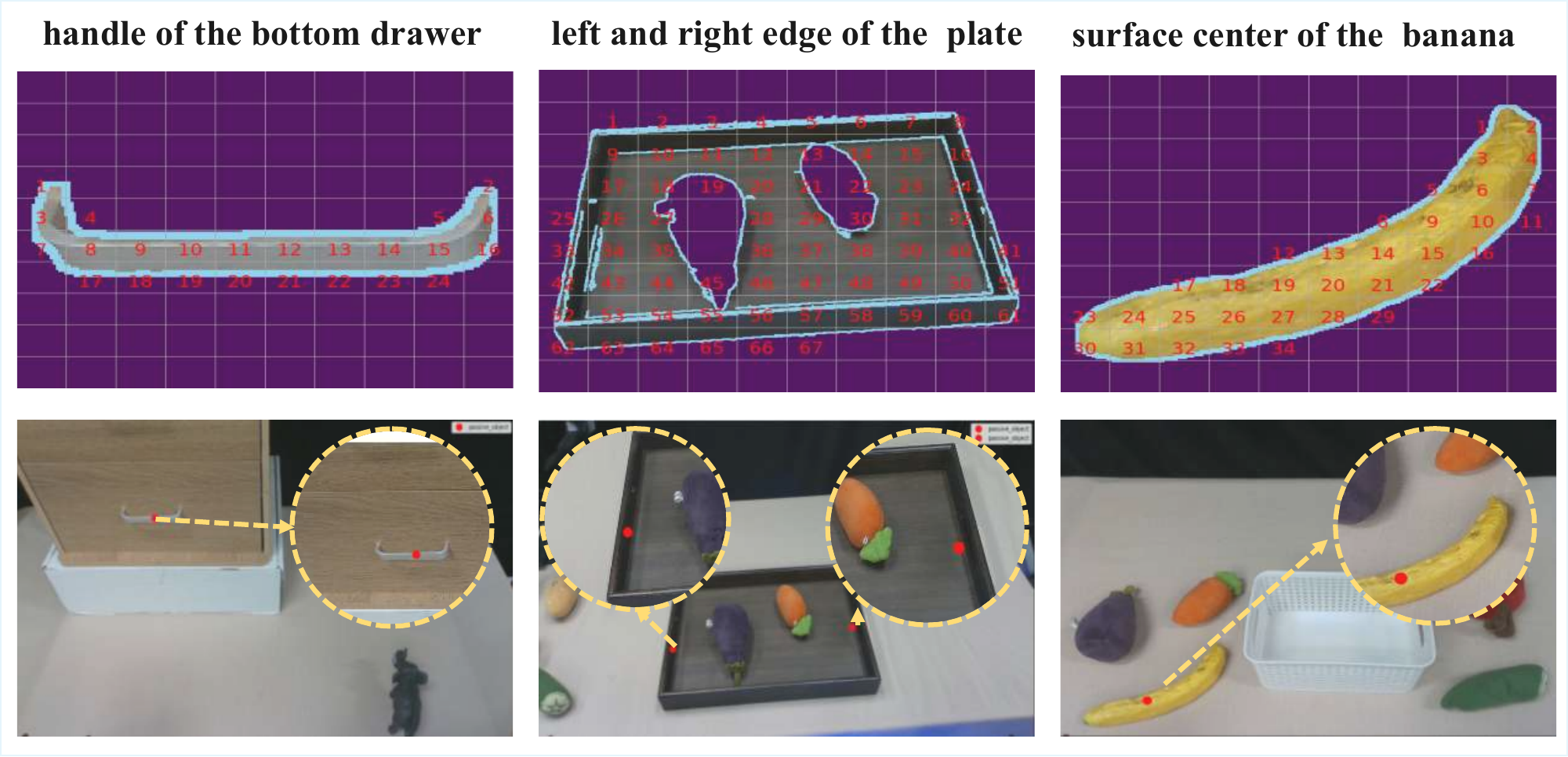}
  \caption{\textbf{Examples of Keypoints Selection.} The top row displays the grid-annotated visual prompts generated for specific objects (handle, plate, banana) after ROI normalization and contour highlighting. The VLM selects specific grid indices (red numbers) based on semantic object part description (e.g., "handle of the bottom drawer", "left and right edge of the plate", "surface center of the banana"). The bottom row illustrates the corresponding 3D manipulation coordinates (red dots) mapped back onto the RGB image, demonstrating the system's ability to translate high-level semantic descriptions into precise physical locations.}
  \label{fig:kpts}
\end{figure*}

\subsection{Fine-grained Keypoint Selection}
Following object detection and segmentation, the system determines the precise 3D manipulation coordinates. We utilize a \textit{Grid-based Visual Prompting} method \cite{ReSem3D} (as shown in Fig. \ref{fig:kpts}) that enables the VLM to perform pixel-level semantic localization without requiring pixel-coordinate output capabilities.

To bridge the gap between semantic instructions (e.g., ``grasp the handle'') and spatial coordinates, we generate a visual prompt for each detected object. This process involves three steps: 

\paragraph{\textbf{ROI Normalization and Enhancement}}
To emphasize the object's geometry, we apply Canny edge detection to the segmented region and highlight the contours in blue. The processed image is then cropped and resized to a standardized resolution to ensure the visibility of the small features and structural details.

\paragraph{\textbf{Grid Overlay}}
A labeled grid (e.g., $10 \times 10$) is overlaid on the object image. To ensure the keypoints fall on the object surface, the grid labels are positioned based on the foreground pixel density within each cell.

\paragraph{\textbf{Semantic Selection}} The VLM receives this grid-annotated image along with the interaction description of the sub-stage. Instead of predicting continuous coordinates, the model performs a classification task by selecting the grid indices that correspond to the desired functional parts.

Once the VLM selects the target grid indices, the system maps the 2D grid centers back to the original image coordinate system $(u, v)$. To obtain the corresponding 3D world coordinates $(x, y, z)$, we employ a \textit{3D Coordinate Extraction} strategy. This mitigates noise in the raw point cloud through \textit{Neighborhood Sampling}, where we sample a local $N \times N$ neighborhood (e.g., $7 \times 7$) around the target pixel, and \textit{Outlier Rejection \& Median Filtering}, which filters out invalid depth values based on spatial continuity constraints. The final 3D coordinate is computed as the spatial median of the valid neighborhood points, ensuring geometric stability for the robotic manipulation.

\begin{figure*}[tbp]
  \centering
  \includegraphics[width=\textwidth]{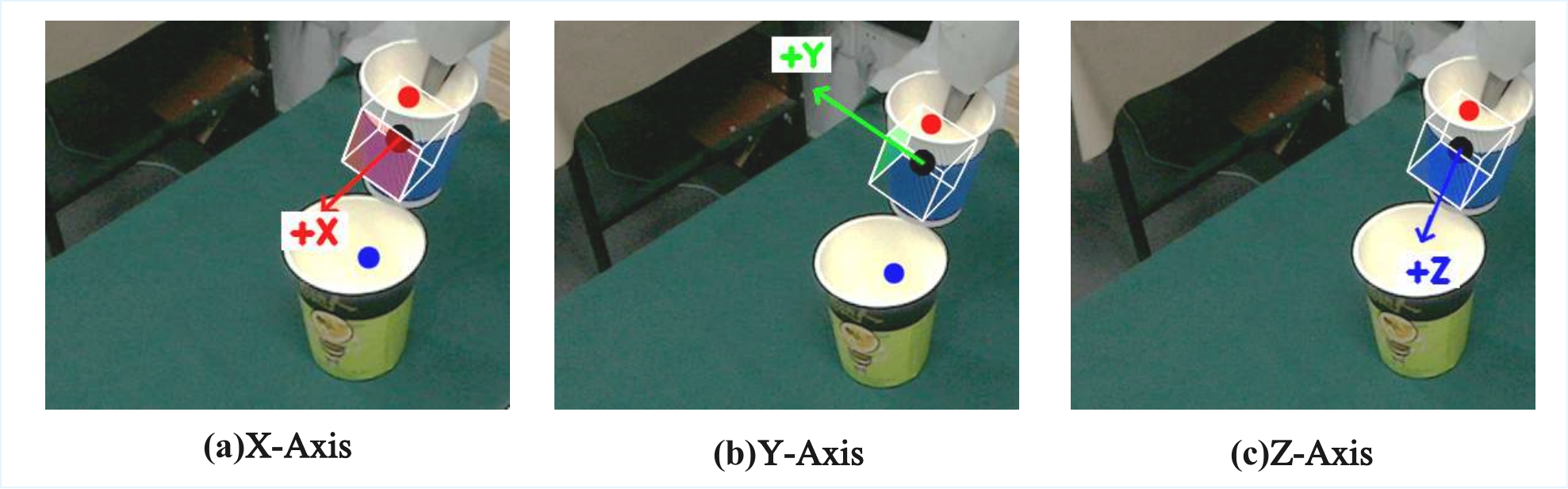}
  \caption{\textbf{Examples of Rotation Annotated Frames.} To ground abstract rotation axes in the physical world, the system projects the end-effector's local coordinate system onto the image. The figure displays the three orthogonal axes provided to the VLM: (a) X-axis (Red), (b) Y-axis (Green), and (c) Z-axis (Blue). The black dot represents the computed geometric median of the target object. These visualizations enable the VLM to intuitively infer the rotation parameters.}
  \label{fig:rotation}
  \vspace{-0.1cm}
\end{figure*}

\subsection{Grasp Pose Generation}
To translate the semantic keypoints into executable 6-DoF grasp poses, we employ a hybrid strategy that adapts to object characteristics and task requirements.

For \textit{Geometric Heuristic Generation}, utilized primarily for small objects (height $\le$ 0.05m), articulated objects and bimanual tasks involving dual keypoints, the system derives grasp poses directly from the geometric properties of the detected bounding boxes (BBoxes). To bridge the perception and manipulation spaces, the four vertices of the BBox's are projected into the robot's base coordinate system. The algorithm determines the grasp orientation by aligning the gripper’s approach vector with the BBox surface normal facing the robot, while aligning the gripper closing direction with the shortest edge of BBox or specific functional edges. Finally, the grasp center is anchored at the task-relevant keypoints which are selected in keypoint selection module.

For general objects, we integrate the AnyGrasp\cite{anygrasp} with our semantic guidance. The process proceeds in two steps:

\paragraph{\textbf{Kinematic Filtering}} Upon retrieving dense grasp candidates, we transform them back to the world frame and apply kinematic constraints. We filter out poses with invalid approaching vectors (e.g., collision-prone orientations) to ensure the reachability of the robotic arm. 

\paragraph{\textbf{Keypoint-guided Selection}} Given the set of candidate grasps $\mathcal{G} = \{g_i\}_{i=1}^N$ generated by the AnyGrasp, where each grasp $g_i$ is parameterized by its rotation $\mathbf{R}_i$, translation $\mathbf{t}_i$, and confidence score $s_i$. Let $\mathbf{p}_{key} \in \mathbb{R}^3$ denote the 3D semantic keypoint generated from keypoint selection module.

We formulate the selection of the optimal grasp $g^*$ as a radius-constrained optimization problem. The goal is to maximize the grasp quality score $s_i$ within a dynamically expanding neighborhood of $\mathbf{p}_{key}$. The optimal grasp is defined as:

\begin{equation}
    g^* = \underset{g_i \in \mathcal{S}_k}{\arg\max} \ s_i
\end{equation}

subject to the spatial feasibility set $\mathcal{S}_k$ at iteration $k$:

\begin{equation}
    \mathcal{S}_k = \{g_i \in \mathcal{G} \mid \|\mathbf{t}_i - \mathbf{p}_{key}\|_2 \le r_k \}
\end{equation}

where the search radius $r_k$ is updated iteratively according to $r_k = r_{init} + k \cdot \Delta r$. The search starts with $k=0$ and increments until $\mathcal{S}_k \neq \emptyset$ or $r_k > r_{max}$. This iterative expansion prioritizes semantic proximity, ensuring that the robot acts upon the specific functional part (e.g., the handle), falling back to a wider search area only if no valid high-quality grasps are available in the immediate vicinity.

\subsection{Rotation Tool}
While translation tasks primarily rely on spatial coordinate perception, many tasks necessitate object rotation to satisfy specific instructions (e.g., ``pour water'' and ``insert a pen''). We propose a VLM-guided framework to determine the rotation parameters. The process consists of two stages:

\paragraph{\textbf{Reference Frame Visualization}} To ground the rotation axes in the physical world, the system first computes the geometric median of the target object's point cloud. It then projects the end-effector's local coordinate axes (X, Y, Z) onto the object's image(As shown in Fig. \ref{fig:rotation}).

\paragraph{\textbf{Parameter Inference}} The VLM receives the current task instruction, the end-effector's rotation matrix and the annotated images from the last step. Then the model predicts rotation parameters: the \textit{Rotation Axis} (local X/Y/Z), \textit{Direction} (clockwise/counter-clockwise), and \textit{Angle} (degrees). This semantic parameterization bridges the gap between high-level language commands and low-level control.

\section{Model Specifications and Prompts Details}

\noindent\textbf{Model Specifications}
In our framework, we primarily utilize the Qwen3 series models\cite{qwen3}. Specifically, the Planner Agent utilizes Qwen3-Max (snapshot \texttt{qwen3-max-\allowbreak2026-01-23}), while the Manipulation and Verification Agents operate on Qwen3-VL-Plus (snapshot \texttt{qwen3\allowbreak-vl-plus-2025-12-19}). Both models are accessed via the Alibaba Cloud DashScope API.

\noindent\textbf{System Prompts}
We list the system prompts  employed in our framework, categorized by their functional roles. Please note that for the sake of brevity and clarity, the prompts presented below are simplified versions. We focus on displaying the core roles, goals, and reasoning constraints.

\textit{a) \textbf{Planning Agent}:}

\begin{promptbox}{1. Multi-Robot Task Planning}
\textbf{Task:} Generate a collaborative JSON plan for a multi-robot system (ArmRobot, SingleArmRobot, MobileRobot) based on \texttt{\#OBSERVATIONS\#}, \texttt{\#ROBOT\_POSITIONS\#}, and \texttt{\#TASK\_GOAL\#}.

\textbf{Reasoning Constraints:}
\textbf{Collaboration:} Prioritize placing objects on tables for transfers (especially for liquids) rather than direct handovers.
\textbf{Dual-Arm Strategy:} For dual-arm robots, combine simultaneous grasps into a single action string (e.g., ``pick A and pick B''), but strictly enforce sequential placement if targeting the same container to avoid collision.
\textbf{Ambiguity:} Set \texttt{further\_allocation} to \texttt{true} only if specific object assignment is unclear among co-located robots.
\textbf{Efficiency:} Schedule independent actions (e.g., navigation vs. manipulation) in parallel steps.

\textbf{Output Format:} JSON array of steps, each containing \texttt{step}, \texttt{parallel}, and \texttt{subtasks} (list of objects with \texttt{robot}, \texttt{action}, \texttt{current\_pos}, \texttt{further\_allocation}, and \texttt{reason}).
\end{promptbox}

\begin{promptbox}{2. Global Failure Recovery}
\textbf{Task:} Generate a new recovery plan to complete the \texttt{\#INSTRUCTION\#} following a runtime error (\texttt{\#ERROR\_MESSAGE\#}). Analyze the \texttt{\#OBSERVATIONS\#}, \texttt{\#ROBOT\_POSITIONS\#}, and the interrupted \texttt{\#INITIAL\_PLAN\#}.

\textbf{Recovery Strategy:} \textbf{Analyze Error:} If the error implies physical inability (e.g., reachability), \textbf{re-allocate} the subtask to a different capable robot (e.g., MobileRobot) and include necessary navigation. \textbf{Execution Continuity:} \textbf{Do not replan} successfully executed operations; resume planning strictly from the failure step.

\textbf{Constraints:} Maintain goal continuity. Apply \textbf{Dual-Arm Optimization} (combined pick, sequential place) where possible. Use \texttt{further\_allocation} only if object ownership is ambiguous.

\textbf{Output Format:} JSON array of steps, each containing \texttt{step}, \texttt{parallel}, and \texttt{subtasks} (list with \texttt{robot}, \texttt{action}, \texttt{current\_pos}, \texttt{reason}, \texttt{further\_allocation}).
\end{promptbox}

\textit{b) \textbf{Manipulation  Agent}:}

\begin{promptbox}{1. Sub-task Planning }
\textbf{Task:} Analyze the \texttt{\#INSTRUCTION\#} and \texttt{\#IMAGE\#} to decompose the task into executable stages \\
\textbf{Skill Library (Primitives):}
\begin{itemize}
    \item grasp\_object / grasp\_object\_with\_two\_arms
    \item place\_object / place\_object\_with\_two\_arms
    \item ...
\end{itemize}
\textbf{Reasoning Constraints:} Operate only on visible objects. Ensure each stage maps to a single atomic skill. Distinguish between \textit{Scenario A} (Robot $\to$ Object) and \textit{Scenario B} (Object $\to$ Object) to correctly assign \texttt{active}/\texttt{passive\_object} fields.

\textbf{Output Format:} JSON list items containing \texttt{stage\_id}, \texttt{dual\_arms}, object names/descriptions, and a concise \texttt{interaction\_description}.
\end{promptbox}

\begin{promptbox}{2. Scene Perception (for \texttt{further\_perception})}
\textbf{Task:} Analyze the \texttt{\#IMAGE\#} and \texttt{\#INSTRUCTION\#} to categorize relevant items into ``Target Objects'' (manipulation targets) and ``Static Objects'' (destinations/containers).

\textbf{Reasoning Logic:} Identify targets with specific attributes and median points (\texttt{point\_2d}). For static objects, \textbf{priority} is given to specific containers over general surfaces (fallback to table only if no container exists).

\textbf{Output Format:} JSON containing \texttt{target\_objects} (list of objects with \texttt{label}, \texttt{point\_2d}) and \texttt{static\_objects} (list of descriptions).
\end{promptbox}

\begin{promptbox}{3. Object Detection }
\textbf{Task:} Identify the \texttt{\#OBJECTS\_STR\#} in the \texttt{\#IMAGE\#} and output their bounding box coordinates and labels.

\textbf{Constraints:} Account for potential partial occlusions (e.g., by the robotic gripper). Ensure the output contains separate items matching the \textbf{exact count} of requested objects, no more, no less.

\textbf{Output Format:} JSON list where each item contains \texttt{label} and \texttt{bbox\_2d}.
\end{promptbox}

\begin{promptbox}{4. Keypoint Selection }
\textbf{Task:} Identify optimal grid labels (red numbers) for ``Active'' and ``Passive'' objects using the provided grid-overlaid \texttt{\#IMAGES\#}(active object(optional), passive objects), \texttt{\#SCENE\_IMAGE\#} and \texttt{\#DESCRIPTIONS\#}.

\textbf{Selection Rules:} Determine if the task is \textbf{bimanual}. For \textbf{tangible} surfaces, select a single optimal label (select 2 only if bimanual). For \textbf{intangible} spaces (e.g., cup center), select symmetrical boundary labels and set \texttt{visible\_and\_tangible} to \texttt{false}. Avoid grids that are too deep inside or on the outermost edges of internal spaces.

\textbf{Output Format:} JSON containing \texttt{bimanual} (bool), \texttt{active\_objects}, and \texttt{passive\_objects} (each with \texttt{name}, \texttt{label}, and \texttt{visible\_and\_tangible}).
\end{promptbox}

\begin{promptbox}{5. Action Primitive Generation}
\textbf{Task:} Decompose a sub\_task into executable Python code using the provided \texttt{\#PLAN\_SO\_FAR\#}, \texttt{\#INSTRUCTION\#}, \texttt{\#KEYPOINTS\#} and \texttt{\#SCENE\_IMAGE\_WITH\_KEYPOINTS\#}.

\textbf{Primitive Library:} Access 
\textbf{Basic Manipulations} (macros: \texttt{grasp}, \texttt{place}, \texttt{lift up}, \texttt{put down}, \texttt{reset home}) and 
\textbf{Spatial Adjustment} (\texttt{move xy}, \texttt{move pose}, \texttt{rotate}, \texttt{align}) and 
\textbf{Constraint interaction} (\texttt{open}, \texttt{close}).

\textbf{Reasoning Logic:} Analyze spatial constraints and object states. Prioritize basic manipulation primitives for standard interactions (e.g., placing on flat surfaces). Use \textbf{Hybrid Solutions} (combining primitives like \texttt{align\_two\_keypoints} + \texttt{rotate}) for complex manipulations.

\textbf{Output Format:} Executable Python code sequence (e.g., \texttt{align\_two\_keypoints(agent, ...)}).
\end{promptbox}

\begin{promptbox}{6. Geometric Rotation Solver}
\textbf{Task:} Determine the optimal rotation parameters (Axis, Direction, Angle) for the robot end-effector to execute the \texttt{\#INSTRUCTION\#}, using the provided \texttt{\#IMAGE\#} and \texttt{\#OBJECT\_FRAME\_VECTORS\#}.

\textbf{Reasoning Logic:} Select the primary functional axis based on object affordance (e.g., perpendicular to spout). Determine direction by analyzing the relative geometry between objects and ensuring liquid safety (directing content away from the gripper).

\textbf{Output Format:} JSON containing \texttt{rotation\_axis} (X/Y/Z), \texttt{rotation\_direction} (positive/negative), and \texttt{angle} (selected from 60, 75, 90, 100).
\end{promptbox}

\begin{promptbox}{7. Sub-task Signature Extractor (for Long-Term Memory)}
\textbf{Task:} Parse the \texttt{\#SUB\_TASK\_DESCRIPTION\#} into a structured semantic signature to facilitate experience recording and retrieval in the experience pool.

\textbf{Parsing Logic:} Identify the \texttt{active object} (initiator) and \texttt{passive objects} (recipients). \textit{Special Rules:} Omit the active object for robot-centric actions (grasp, pick, open, close) where the agent is implicit. Exclude specific positional descriptions (e.g., "at Position X") to ensure the output serves as a generalizable template.

\textbf{Output Format:} JSON containing \texttt{action}, \texttt{active\_object}, and a list of \texttt{passive\_objects}.
\end{promptbox}

\textit{b) \textbf{Verification Agent}:}

\begin{promptbox}{1. Task Verification }
\textbf{Task:} Analyze the \texttt{\#INSTRUCTION\#} and pre/post-action images (\texttt{\#Image\_PRE\#}, \texttt{\#Image\_POST\#}) to assess if the robot successfully executed the command.

\textbf{Evaluation Rules:} Verify action completion and object states. Identify \textbf{gripper positions}. 
For \textbf{rotational tasks}, strict precision is not required. For dual-arm systems, success is valid if either gripper completes the target.

\textbf{Output Format:} JSON object: \texttt{\{"is\_success": true\}} or \texttt{\{"is\_success": false, "reason": "failure explanation"\}}.
\end{promptbox}

\begin{promptbox}{2. Local Failure Recovery }
\textbf{Task:} Analyze pre/post-action images (\texttt{\#Image\_PRE\#}, \texttt{\#Image\_POST\#}) to verify the \texttt{\#CURRENT\_OPERATION\#}. If failed, generate a recovery sequence based on the execution \texttt{\#HISTORY\#}.

\textbf{Recovery Logic:} Compare states to determine outcome. If \textbf{Condition Loss} occurs (e.g., object dropped), retrieve the preparatory step from history and append the current step. If \textbf{Ineffective Action} occurs (state unchanged), strictly retry the current operation. All steps must match exact strings from history.

\textbf{Output Format:} JSON containing an ordered list of \texttt{replan\_steps}.
\end{promptbox}

\clearpage